%% file: main.tex
\documentclass{article} 
\usepackage{iclr2024_conference,times}

\input{math_commands.tex}

\usepackage{hyperref}
\usepackage{url}
\usepackage{booktabs}
\usepackage[group-separator={,},output-decimal-marker={.}]{siunitx}

\usepackage[textsize=tiny,textwidth=3cm]{todonotes}

\graphicspath{{graphs/}}  

\title{Representation Learning of Daily Movement \\ Data Using Text Encoders}

\author{Alexander Capstick \\
Imperial College London\\
\texttt{alexander.capstick19@imperial.ac.uk} \\
\And
Tianyu Cui \\ Imperial College London \\ \texttt{t.cui23@imperial.ac.uk}
\And
Yu Chen \\ Imperial College London \\ \texttt{yu.chen@imperial.ac.uk}
\And
Payam Barnaghi \\  Imperial College London  \\ \texttt{p.barnaghi@imperial.ac.uk}
}

\iclrfinalcopy 
\begin{document}

\maketitle

\begin{abstract}
Time-series representation learning is a key area of research for remote healthcare monitoring applications. In this work, we focus on a dataset of recordings of in-home activity from people living with Dementia. We design a representation learning method based on converting activity to text strings that can be encoded using a language model fine-tuned to transform data from the same participants within a $30$-day window to similar embeddings in the vector space. This allows for clustering and vector searching over participants and days, and the identification of activity deviations to aid with personalised delivery of care.
\end{abstract}

\section{Introduction}

In remote healthcare monitoring applications, time-series data is continuously collected using wearables or Internet of Things devices. This scale of data collection necessitates machine learning techniques for analysis and delivery of care. Additionally, time-series data is challenging to label \citep{ching2018opportunities}, requiring representation learning through self-supervised, semi-supervised, or unsupervised methods to extract insights. 
This transforms data into useful encodings for downstream tasks and analysis and can often be performed on datasets with few or no labels \citep{franceschi2019unsupervised, eldele2021time, zerveas2021transformer, yang2022unsupervised}. 

Considering a neural network as a feature extractor followed by a predictor \citep{murphy2022probabilistic}, supervised learning becomes joint representation learning and predictive modelling. This fact inspires most of the self-supervised or semi-supervised representation learning literature to date since predictive tasks can be defined using the time-series itself, using techniques such as data masking, shuffling, or contrastive learning \citep{balestriero2023cookbook}.

In this work, we focus on collections of time-series containing irregularly appearing discrete values. These could be time-series of electronic healthcare records or remote monitoring activity. 
We present initial results of a representation learning method that benefits from advancements in large language modelling, text encoding, and vector search to aid in clustering, finding similar clinical cases, identification of changes in data patterns, and to support personalised care delivery. In this way, our method utilises a language model's pre-trained representations of different in-home locations which we hypothesise allow for a more meaningful encoding of data.

\subsection{Background}


In representation learning, the goal is to encode raw data as vectors that are useful for further analysis, usually by harnessing deep learning. The literature on this topic is broad, including theoretical techniques concerned with few-shot prediction \citep{wang2020generalizing}, continual learning \citep{hadsell2020embracing}, multimodal learning \citep{baltrusaitis2019multimodal, xu2023multimodal}, and applications to healthcare \citep{alanazi2022using}, language modelling \citep{vaswani2017attention}, and image generation \citep{radford2021learning}. In many ways, all neural networks utilise representation learning.

For time-series data, varied methods of learning data representations have been proposed. For example, \citet{franceschi2019unsupervised}, inspired by \citet{mikolov2013distributed}, train an encoder based model using a triplet loss by selecting random subsets of the same time-series recording as similar, and random subsets of other recordings as dissimilar; \citet{eldele2021time} propose training models to maximise the similarity between embeddings of weak and strong augmentations of data, which they show are useful for downstream tasks; and \citet{zerveas2021transformer} use a transformer-encoder architecture \citep{vaswani2017attention} and a linear projection over a time-series to learn representations by masking inputs and predicting the hidden values.


Large language models (LLMs), based on self-attention \citep{vaswani2017attention}, are examples of time-series representation learning models that have found significant success due to their scalability and parallel design \citep{lin2022surveytransformers}. Fine-tuning these models have allowed for the transfer of learning to specialised tasks initially outside of their intended uses \citep{brown2020language, wei2021finetuned, agrawal2022large, ding2023parameter}. Additionally, since language models are trained on billions of text documents from varied domains, they have been shown to encode relational knowledge \citep{petroni2019language} of text without being explicitly trained for these tasks.

Progress has been made in adapting LLMs for use outside of language modelling. \citet{gruver2023large} experiment with GPT-3 \citep{brown2020language} and LLaMA-2 \citep{touvron2023llama}, finding they have a surprisingly good ability to zero-shot forecast time-series encoded as strings of numerical digits. Moreover, \citet{jin2023time} use prompt engineering and reprogramming to adapt LLMs for classification tasks. In both, the authors test numerical time-series data focused on predictive tasks.

\subsection{Our contribution}

Focusing on in-home activity data collected passively using low-cost sensors from the homes of people with Dementia, we learn representations of time-series for tasks such as activity clustering, searching for similar participants and patterns, and measuring deviations in activity. This technology would aid clinical teams in personalised care delivery planning or assessing risks to morbidities by referencing similar participants with known health trajectories and care requirements. Appendix \ref{sec:reproducibility} discusses the availability of code and data for the reproduction of this work.

\section{Methods}

\subsection{Language Modelling}
\label{sec:language_modelling}

Following \citet{gruver2023large}, language models are typically auto-regressive models trained on a collection of sequences, $\mathcal{U} = \{U_1, ..., U_N \}$, where $U_i = (u_1, ..., u_{n_i})$ and each $u_i$ is a token representing a single piece of text from a vocabulary $\mathcal{V}$. Since each predicted token is dependent on all previous tokens in the sequence, for a model parameterised by $\theta$, we have $\log p(U_i | \theta) = \sum_{j=1}^{n_i} \log p(u_j | u_{0:j-1}, \theta)$. The model parameters, $\theta$ are learnt by maximising this likelihood over the entire dataset (i.e, jointly over $\mathcal{U}$).
The tokeniser component of a language model encodes an input string into a sequence of tokens from $\mathcal{V}$. 

The transformer architecture \citep{vaswani2017attention}, including an encoder and decoder, is a popular choice for language modelling. BERT \citep{devlin2019bert}, based on the transformer encoder architecture only, is trained with $\log p(U_i | \theta) = \sum_{j=1}^{n_i} \log p(u_j | u_{-j}, \theta)$ (with $u_{-j}$ as all tokens except $u_{j}$) and designed to be fine-tuned \citep{murphy2022probabilistic}. This model outputs an embedding for each input token, contextualising them within a sequence. 

\citet{reimers2019sentencebert} apply mean-pooling to token embeddings to produce sentence embeddings; a single vector (of length given by the token embedding length) representing an entire sequence (with a maximum length defined by the BERT architecture). In parallel, \citet{wang2020minilm} present a method for distilling a BERT-like model into smaller models useful for fine-tuning. The authors make this distilled model, MiniLM (and subsequently MiniLM V2) publicly available. These works lead to a sentence embedding model based on a fine-tuned MiniLM V2 model \citep{reimers2019sentencebert, wang2020minilm, wolf2020transformers} referred to here as ``SE-MiniLM".

Sequence (or sentence) embedding models can be fine-tuned using a triplet loss \citep{hoffer2015deep}, measuring the cosine similarity between the embeddings of an anchor sequence and a similar and dissimilar sequence, such that similar sequences have similar embeddings.

Semantic vector search uses the learnt embeddings of objects to find similar objects. This can be done using various similarity measures. In this work, we use cosine similarity.

\subsection{The dataset}

We have access to a dataset collected from $134$ people living with Dementia, containing in-home data recorded between 2021-07-01 and 2024-01-30. Each participant in this study has passive infrared (PIR) sensors positioned around their house, allowing us to collect for each day, the times that locations in the home were visited by occupants. Additionally, the dataset contains data collected from sleep mats that record the time the person living with Dementia enters or leaves their bed. On average, the number of days recorded per participant is $492$ (a minimum of $5$, median of $513$, and maximum of $943$) and in total, this dataset contains \num{65962} recorded days. Further information is given in Appendix \ref{sec:further_dataset_information}.

\subsection{Our method}

\begin{figure}[t]
    \centering
    \includegraphics[width=1\linewidth]{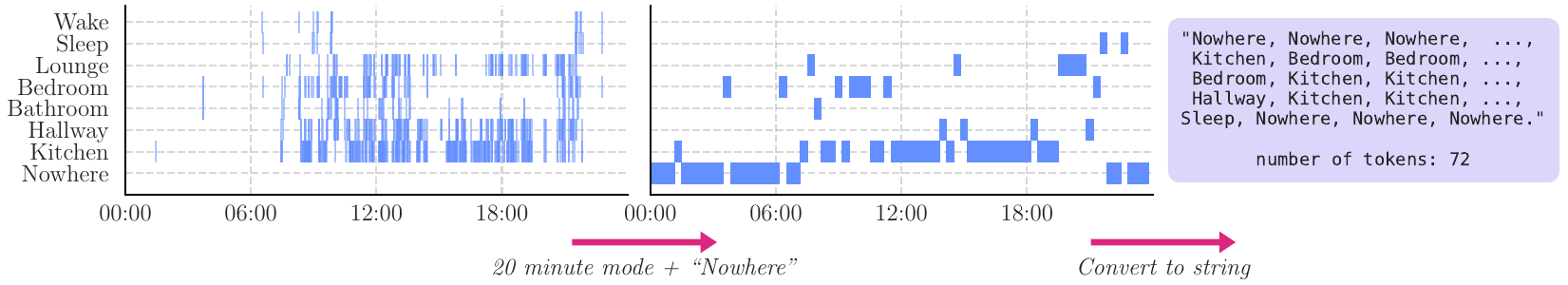}
    \caption{\textbf{Preprocessing of data.} For a single participant, we present their data collected over a single day. The left graph shows the raw measurements; the central graph shows the measurements after the mode of each 20 minute window is taken and a token for no activity is assigned (``Nowhere"); and the right graph shows the day as a single text string that can be interpreted by a language model.}
    \label{fig:single_day_activity}
\end{figure}

This work presents initial results from a system capable of encoding days of recorded activity data into vectors that are useful for care delivery. We transform each participant's daily time-series of events into strings of text that are encoded using a sentence embedding model (Figure \ref{fig:single_day_activity}). 

We first transform each day into a fixed length vector by using the most recorded (modal) location for each $20$ minute window and by assigning a a token, ``Nowhere", to all windows without any sensor readings. This is then transformed into a text string. The SE-MiniLM model described in Section \ref{sec:language_modelling} is fine-tuned on this collection of day-strings using a triplet loss and by assigning a similar and dissimilar example for every string. Here, we make use of the pre-trained word embeddings' learnt semantic relationship between locations which we hypothesise allows for more meaningful encoding of activity. Inspired by \citet{franceschi2019unsupervised}, for a given day-string we uniformly sample another string generated from the same participant within a $30$-day window and label it as similar, and uniformly sample a string from a different participant to label as dissimilar. 
In this way, we are training the model to learn personalised representations of day-strings that allow for the detection of activity changes.
Further details are available in Appendix \ref{sec:method_further_details}.

Daily embeddings can now be clustered using unsupervised techniques, searched over using vector searches, or similarity measured over time to estimate changes in behaviour.

\section{Experiments}
\label{sec:experiments}

\begin{figure}[t]
    \centering
    \includegraphics[width=1\linewidth]{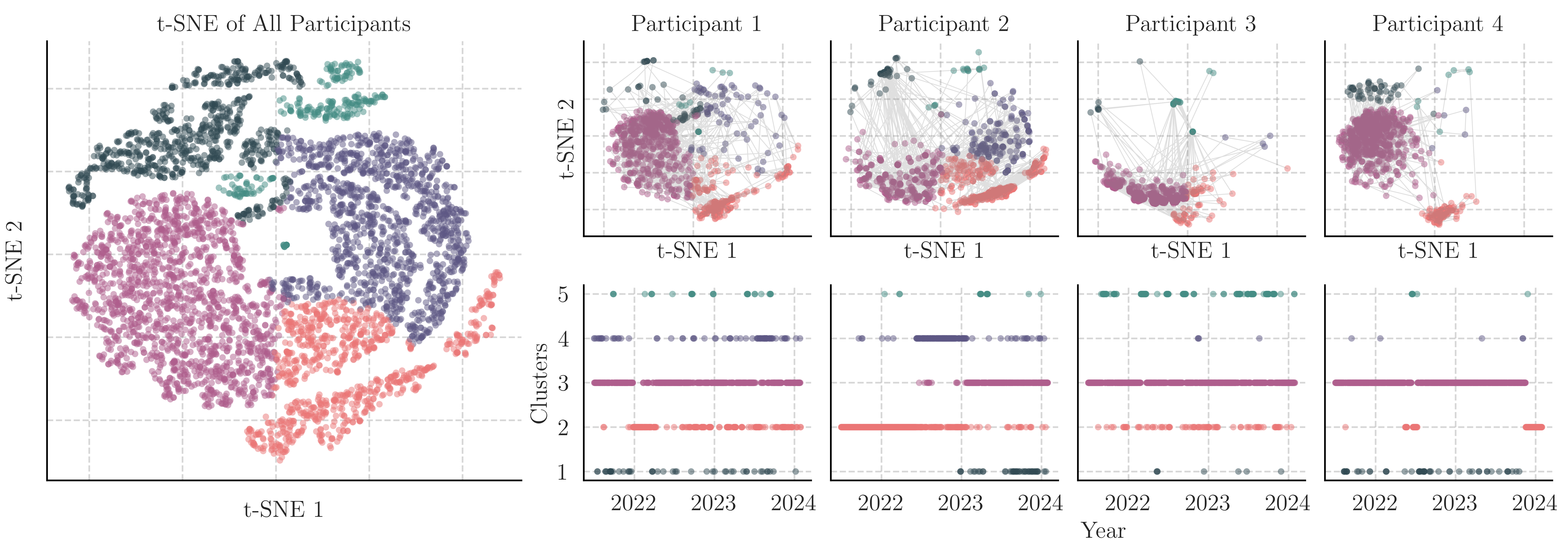}
    \caption{\textbf{t-SNE of embeddings.} The left plot shows the t-SNE transformation of \num{5000} day-string embeddings, coloured by their cluster value. On the upper row of the right hand figures, we show the t-SNE embeddings for the $4$ participants with the most data, where a line indicates consecutive days. On the lower row, we show the daily cluster values in time for the same participants.}
    \label{fig:tsne_clusters}
\end{figure}

After using $k$-means to cluster the learnt embeddings, we found $5$ clusters (Appendix \ref{sec:further_experiments_kmeans}), revealing $5$ categories of days. Figure \ref{fig:tsne_clusters} shows these clusters after embeddings are transformed using t-SNE \citep{lauren2008visualizings}. By viewing these t-SNE embeddings for different participants (on the upper right-hand side), we can visualise $4$ participants' journeys in the embedding space (further participants are visualised in Appendix \ref{sec:further_tsne}). The lower right-hand plot shows the cluster value for each day of data for $4$ participants, which tells us how frequently different participants move between clusters or when activity changes. By exploring with clinicians what these clusters represent semantically, we can gain insight into participants' behaviour and care requirements over time. For example, we found that cluster $1$ is characteristic of small amounts of recorded activity (Appendix \ref{sec:semantic_clustering}) and potentially adverse health conditions. In Appendix \ref{sec:further_clustering_proportions}, we visualise the proportion of participants in each cluster over time, gaining an insight into the behaviour of the cohort as a whole.

Using the cosine similarity between embeddings, for each day of data we can identify the most similar other days (Appendix \ref{sec:vector_search}). We can also use this to find the days that are most similar to those labelled with adverse health conditions. For example, since we know the days that participants were tested for a urinary tract infection (UTI), we can calculate the similarity between days labelled as positive and negative for each participant. We find that the mean ($\pm$ standard deviation) of the average intra-participant cosine similarity between positive and negative days of UTI is $0.53$ ($ \pm 0.36$) and $0.39$ ($ \pm 0.50$) respectively, suggesting that for each participant, positive days of UTI are more similar than to negative days of UTI (Appendix \ref{sec:uti_similarity_information}).

\begin{figure}[t]
    \centering
    \includegraphics[width=1\linewidth]{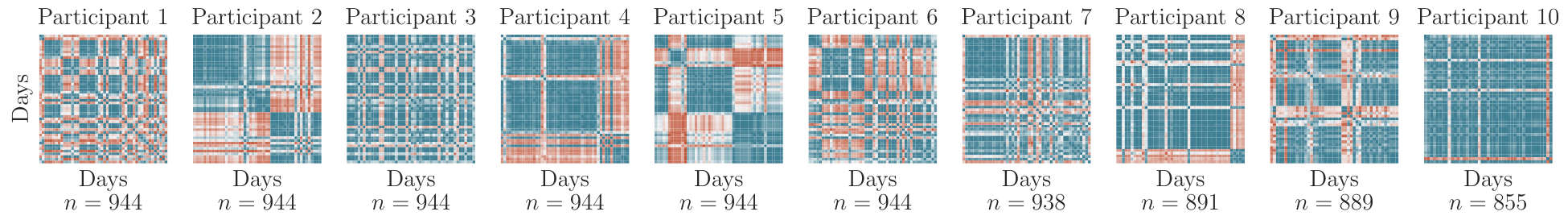}
    \caption{\textbf{Day similarity.} Each image shows the cosine similarity between every $20$th recorded day for $10$ participants. Blue, white, and red correspond to a cosine similarity of $1$, $0$, and $-1$ respectively. The range of the days are represented by each axis of the plots and is given by $n$.}
    \label{fig:patient_similarity}
\end{figure}

Furthermore, we can measure the cosine similarity between each day and each participant to visualise how their activity varies. This is shown in Figure \ref{fig:patient_similarity}, which reveals participants with frequent changes in home activity and participants with more consistent activity. For example, participants $1$ and $6$ have frequent changes in daily activity with small regions of consistency in their activity. However, participant 2 had very consistent activity that switched suddenly to a different state of consistent activity, suggesting a large change in daily routines. Visualising intra-participant similarity, as we have here, could notify healthcare monitoring teams of changes in behaviour that might require changes in care delivery or further investigation.

There are a few directions for future research that are worth highlighting. 
Firstly, we would like to explore other ways of converting our days of activity to text strings and avoid the aggregation of data using the mode (shown in Figure \ref{fig:single_day_activity}), and evaluate how synonyms of the location tokens might change the learnt representations (for example, if ``lounge" is replaced with ``living room"). 
By experimenting with synonyms of the locations used during training, we could gain insights into how a fine-tuned model could be used in differing environments (for instance, we could explore whether a model fine-tuned on in-home data generates meaningful representations of data collected from an assisted living environment).
Further, it would be useful to experiment with different assumptions for generating positive and negative samples when fine-tuning the language model and explore the limitations in the current assumptions. 
We could also consider whether a state space model \citep{gu2023mamba}, which avoids requiring a maximum sequence length, might be a useful replacement for the BERT-based model presented in this work. 
Additionally, we would like to explore alternative methods for clustering that involve domain knowledge from clinicians to improve the utility of activity clustering, and further understand the use-case for identifying similar days to those labelled with adverse health conditions. 
To improve our evaluation of the learnt representations, we aim to test their usefulness in a series of classification tasks to gain additional insights into the information they encode about a participant and their activity, as well as compare the results with methods not relying on pre-trained embeddings or language models.

\section{Conclusion}

This work presents the initial results of a discrete-valued time-series representation learning method, focused on embedding days of in-home activity data, and with applications to other domains. We show that the learnt embeddings are useful for clustering, vector search, and measuring behavioural change. We believe this method shows promise, and we look forward to exploring this further.

\section*{Author contributions}

\textbf{AC:} Conceptualisation, Methodology, Software, Data Processing, Investigation, Writing - Original Draft, Review and Editing, Visualisation;
\textbf{TC, YC:} Writing - Review and Editing; 
\textbf{PB:} Conceptualisation, Methodology, Writing - Original Draft, Review and Editing, Supervision, Funding Acquisition.

\section*{Acknowledgments}

This study is funded by the UK Dementia Research Institute (UKDRI) Care Research and Technology Centre funded by the Medical Research Council (MRC), Alzheimer's Research UK, Alzheimer’s Society (grant number: UKDRI–7002), and the UKRI Engineering and Physical Sciences Research Council (EPSRC) PROTECT Project (grant number: EP/W031892/1). Infrastructure support for this research was provided by the NIHR Imperial Biomedical Research Centre (BRC) and the UKRI Medical Research Council (MRC).
The funders were not involved in the study design, data collection, data analysis or writing the manuscript.


\bibliography{main}
\bibliographystyle{iclr2024_conference}

\newpage

\appendix
\section{Appendix}

\subsection{Availability of the dataset and code}
\label{sec:reproducibility}

The code to fine-tune the model presented in this work is made available\footnote{\url{https://github.com/alexcapstick/Text-Encoders-For-Daily-Movement-Data}}. The dataset and IPython notebook used to generate the figures will not be made publicly available due to their sensitive nature.

Experiments were conducted using python $3.11.5$, torch $2.1.0$ \citep{paszke2019pytorch}, transformers $4.34.1$ \citep{wolf2020transformers}, sentence-transformers $2.4.0$ \citep{reimers2019sentencebert}, scikit-learn $1.3.2$ \citep{scikitlearn}, numpy 1.26.1 \citep{harris2020array}, and pandas $2.1.2$ \citep{reback2020pandas}.

\subsection{Further information on the dataset}
\label{sec:further_dataset_information}

Within this work, we evaluate methods on a dataset containing sensor recordings from the homes of people living with Dementia.

Passive infrared (PIR) sensors are installed at multiple locations within the homes of people living with Dementia, and a sleep mat is positioned under the mattress of the person with Dementia. The PIR sensors can detect motion within $9$ metres and with a maximum angle of $45^\circ$ and the sleep mat device can monitor when the participant enters or exits the bed. Within this work, we focused on PIR sensors located in the Lounge, Kitchen, Hallway, Bedroom, and Bathroom.

\begin{figure}[ht]
    \centering
    \includegraphics[width=1\linewidth]{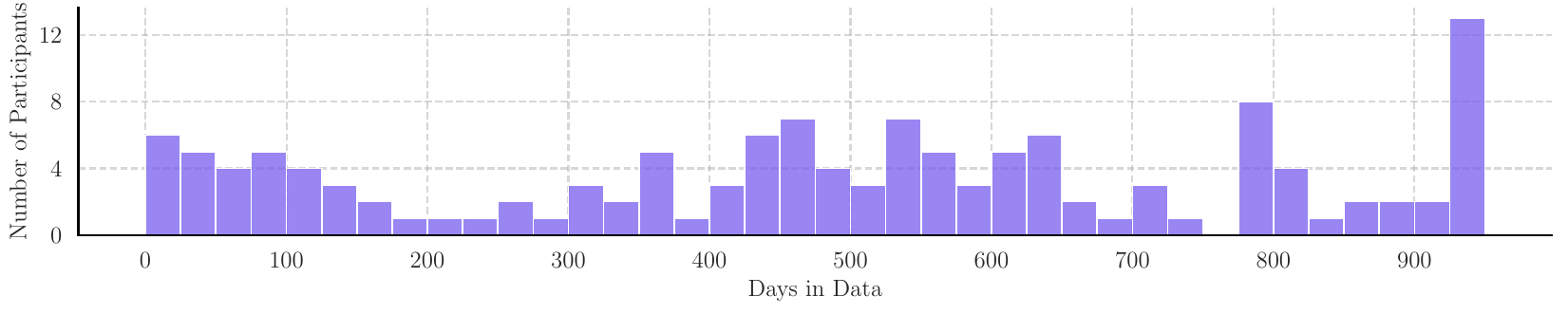}
    \caption{\textbf{Number of days recorded.} Histogram showing the number of participants with a given number of days of data recorded.}
    \label{fig:n_days_histogram}
\end{figure}

\begin{figure}[ht]
    \centering
    \includegraphics[width=1\linewidth]{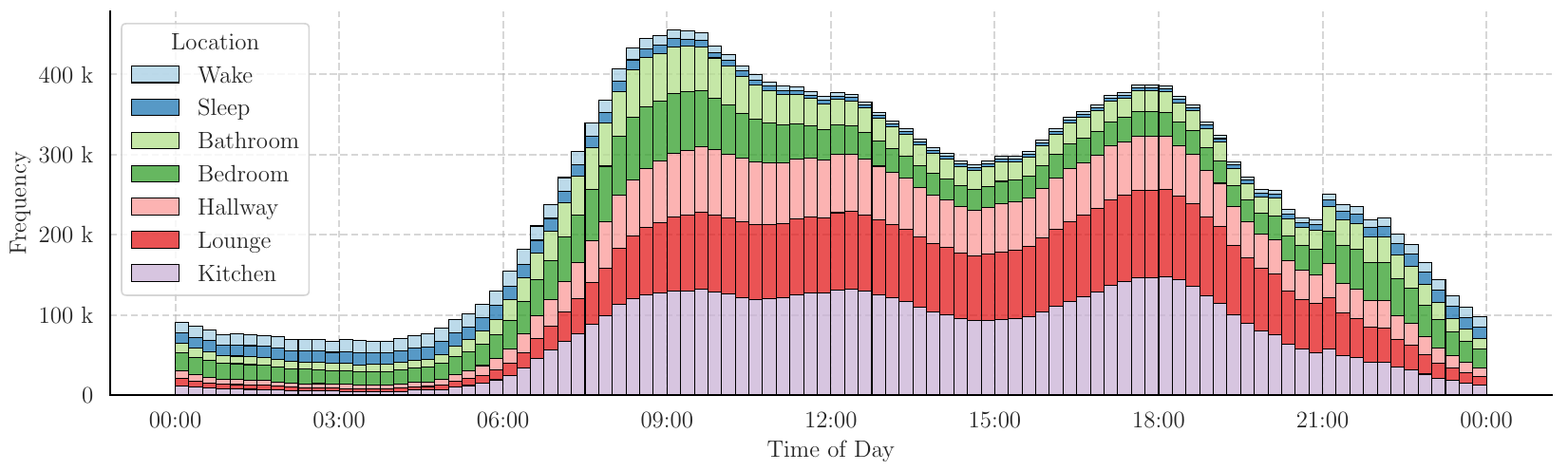}
    \caption{\textbf{Time of location recordings.} Histogram showing the number of sensor recordings by time of day.}
    \label{fig:location_histogram}
\end{figure}

We analyse data collected between 2021-07-01 and 2024-01-30, containing \num{65962} days from $134$ participants. The average number of days recorded per participant is $492$ (with a minimum of $5$, median of $513$, and maximum of $943$). Figure \ref{fig:n_days_histogram} shows the distribution of the number of days recorded from each person living with Dementia, and \ref{fig:location_histogram} shows the sensor recordings by time of day, illustrating the richness of the dataset.

Each data point contains the participant ID, the timestamp of recording (to the nearest second), and the location at which activity was detected or whether the sleep mat recorded someone entering or exiting the bed. This amounted to \num{24467307} data points, shown in Figure \ref{fig:location_histogram}.

As part of the data collection, a subset of participants opted to regularly provide urine samples, which were analysed for suspicion of urinary tract infection. To calculate the similarity between labelled days presented in Section \ref{sec:experiments}, we required participants to have at least $1$ negative result and $2$ positive results (discussed further in Appendix \ref{sec:uti_similarity_information}). This left us with $154$ days labelled as UTI positive ($77$ days)  or negative ($77$ days).

\subsection{Further details of our method}
\label{sec:method_further_details}

For each participant and day, we have access to the sensor recordings to the nearest second. As the tested language model is based on a predefined transformer architecture, it requires an input sequence of less than $256$ tokens in length. Due to the irregular frequency of our time-series, we must aggregate recordings so that they have a fixed length. We do this by calculating the modal sensor reading for each $20$ minute period (and uniformly sample from these modes when more than one exists), which transforms each day into a set of tokens of length $72$ (which is a discrete-valued and regular-frequency time-series). This time-series is then converted into a text string, which can be processed by a language model. 
Using a pre-trained model allows us to make use of the differences in the meaning of, for example, ``Lounge", ``Bedroom", and ``Kitchen" and the literature surrounding sentence embeddings.

These text strings are processed by SE-MiniLM, a sentence embedding model based on a fine-tuned MiniLM V2 model (which itself was pre-trained) \citep{reimers2019sentencebert, wang2020minilm, wolf2020transformers} \footnote{\url{https://huggingface.co/sentence-transformers/all-MiniLM-L12-v2}}. We further fine-tune this model by labelling day-strings produced within $30$ days of each other by the same participant as similar and day-strings produced by different participants as dissimilar. This was done to learn differences between long-term changes in behaviour, and to group participants with similar daily routines into similar regions in the vector space defined by the sentence embeddings.

To fine-tune SE-MiniLM, we randomly sample triplets of text strings from the processed daily activity: one query sentence, one similar sentence, and one dissimilar sentence. For each epoch, we randomly sample \num{100000} triplets from the dataset and assign them to batches of size $256$. We evaluate the sentence embeddings using a triplet loss applied over each triplet, which was optimised using the AdamW algorithm \citep{loshchilov2019decoupled} with a learning rate of $2\times 10^{-5}$ and a weight decay of $0.01$. A linear warm-up learning rate scheduler was used with a warm-up of \num{10000} steps.

\subsection{Further clustering results}
\label{sec:further_experiments_kmeans}

\begin{table}[ht]
\centering
\label{table:clustering}
\caption{\textbf{Silhouette score for $k$-means clustering on the base and fine-tuned models.}}
\begin{tabular}{cccccccccc}
\toprule
Fine-tuning & 2 & 3 & 4 & 5 & 6 & 7 & 8 & 9 & 10 \\
\midrule
True & 0.39 & 0.47 & 0.51 & 0.57 & 0.54 & 0.53 & 0.52 & 0.52 & 0.51 \\
False & 0.42 & 0.21 & 0.18 & 0.16 & 0.16 & 0.14 & 0.14 & 0.17 & 0.16 \\
\bottomrule
\end{tabular}
\end{table}

$k$-means is used to cluster the embedded days of data. We use the $k$-means $++$ algorithm \citep{arthur2007kmeans} with a euclidean distance metric, and numbers of clusters varying from $2$ to $10$. Table \ref{table:clustering} shows the silhouette scores of this experiment, in which $5$ is chosen as optimal.

Embeddings are normalised using the mean and standard deviation of each feature before clustering is performed.

\subsection{Brief exploration of the meaning of clusters}
\label{sec:semantic_clustering}

\begin{figure}[ht]
    \centering
    \includegraphics[width=1\linewidth]{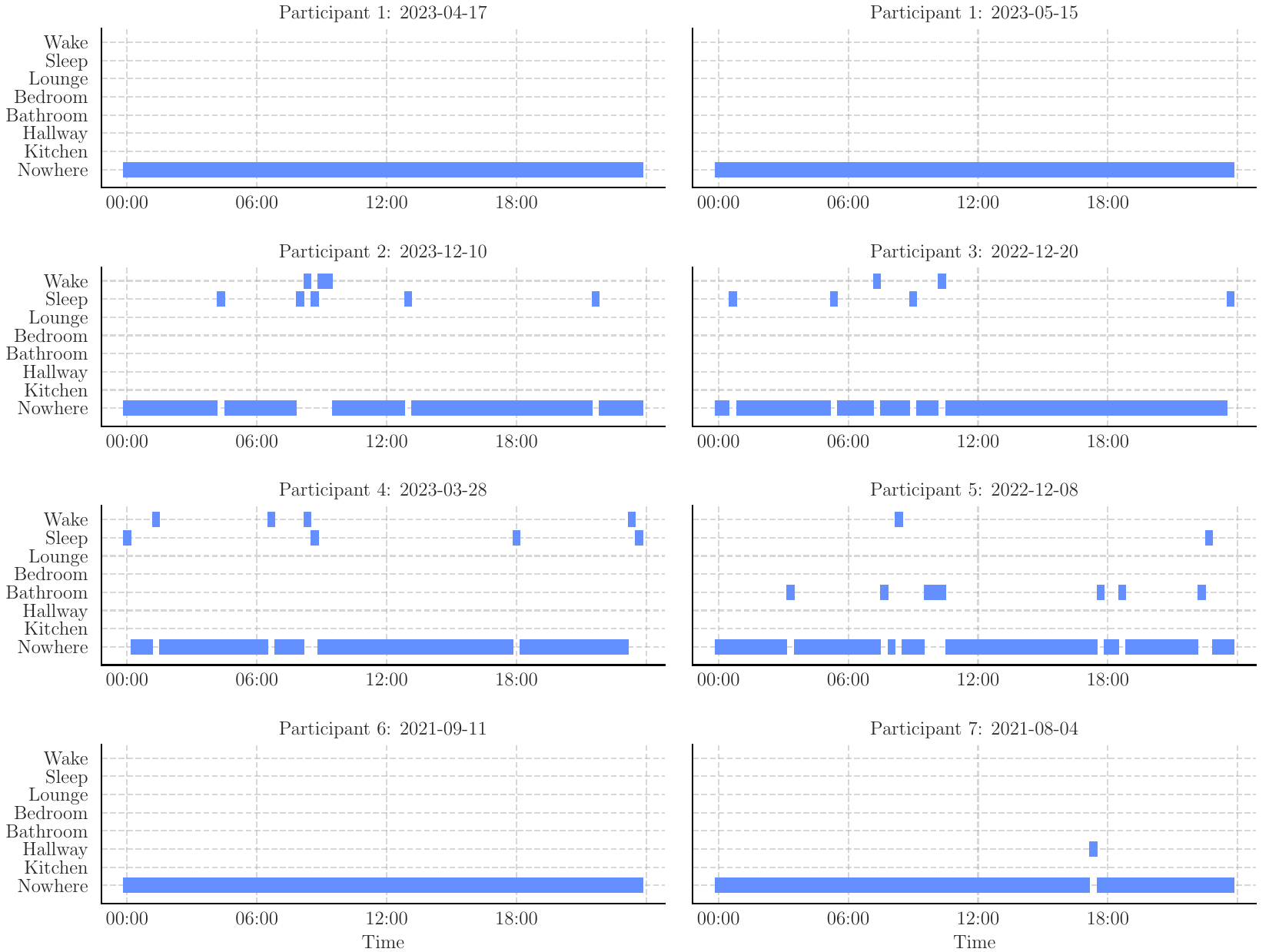}
    \caption{\textbf{Days of data in cluster 5.} We present $8$ randomly chosen days of data from cluster $5$, which is characteristic of sparse recorded activity.}
    \label{fig:cluster_5}
\end{figure}

In Figure, \ref{fig:cluster_5}, we show days of data in cluster $5$, which correspond to days with low levels of recorded activity, which are likely due to sensor recording failures, participants being away from home, or days with little in-home activity (possibly due to adverse health conditions).

\begin{figure}[ht]
    \centering
    \includegraphics[width=1\linewidth]{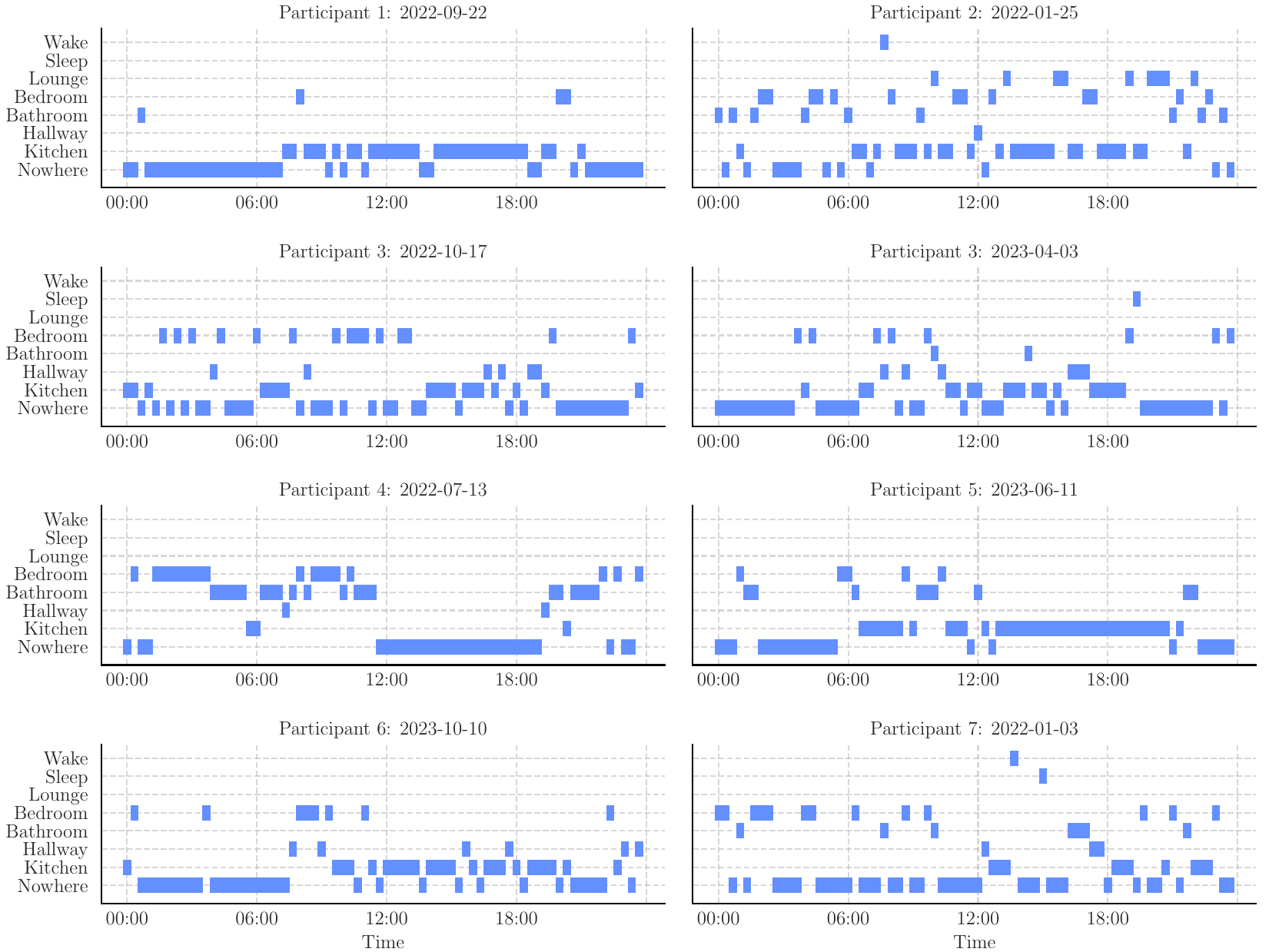}
    \caption{\textbf{Days of data in cluster 1.} We present $8$ randomly chosen days of data from cluster $1$, which is characteristic of frequent bedroom and kitchen activity.}
    \label{fig:cluster_1}
\end{figure}

Similarly, Figure \ref{fig:cluster_1} shows $8$ random days from cluster $1$, which could represent those days with high amounts of kitchen or bedroom activity. Although this cluster could be used to semantically classify a day, it does not seem to have meaningful clinical utility, suggesting more sophisticated methods for learning clusters are required.

\subsection{Further t-SNE plots}
\label{sec:further_tsne}

To visualise the embedding space in two dimensions, we used t-SNE \cite{lauren2008visualizings}, a common dimension reduction method for visualisation. We set the perplexity to $30$, early exaggeration factor to $12$, and the learning rate to $1377$ (number of samples divided by $48$) \citep{scikitlearn}.

Embeddings are normalised using the mean and standard deviation of each feature before applying t-SNE.

\begin{figure}[ht]
    \centering
    \includegraphics[width=0.8\linewidth]{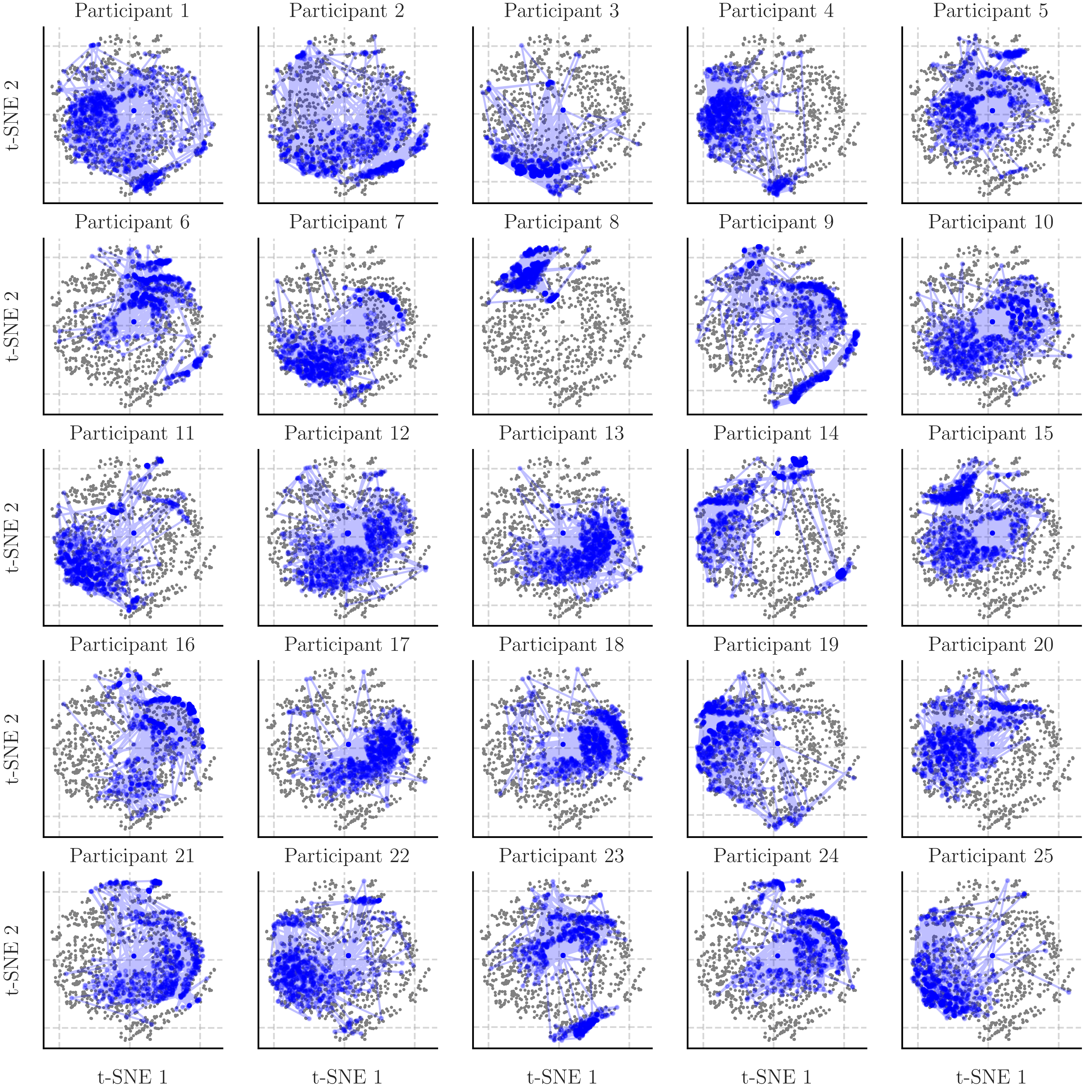}
    \caption{\textbf{t-SNE of embeddings and journeys.} The t-SNE embeddings of \num{1000} randomly sampled days of data (in grey) with the embeddings of a given participant's data overlayed (in blue) with consecutive days connected by a line. Here, we show the $25$ participants with the most recorded days of data.}
    \label{fig:tsne_patient_journey_some}
\end{figure}

In Figure \ref{fig:tsne_patient_journey_some} we present some additional participants and their journeys in the embedding space. Here we plot the $25$ participants with the most recorded days of data. Notice that some participants' activity is significantly more localised than others, suggesting less variability in their activity day-to-day. We also observe some participants that have similar journeys in the transformed embedding space. Participants $15$ and $20$, for example, occupy similar areas in the t-SNE transformed vector space.

\subsection{Clustering over time}
\label{sec:further_clustering_proportions}

\begin{figure}[ht]
    \centering
    \includegraphics[width=1\linewidth]{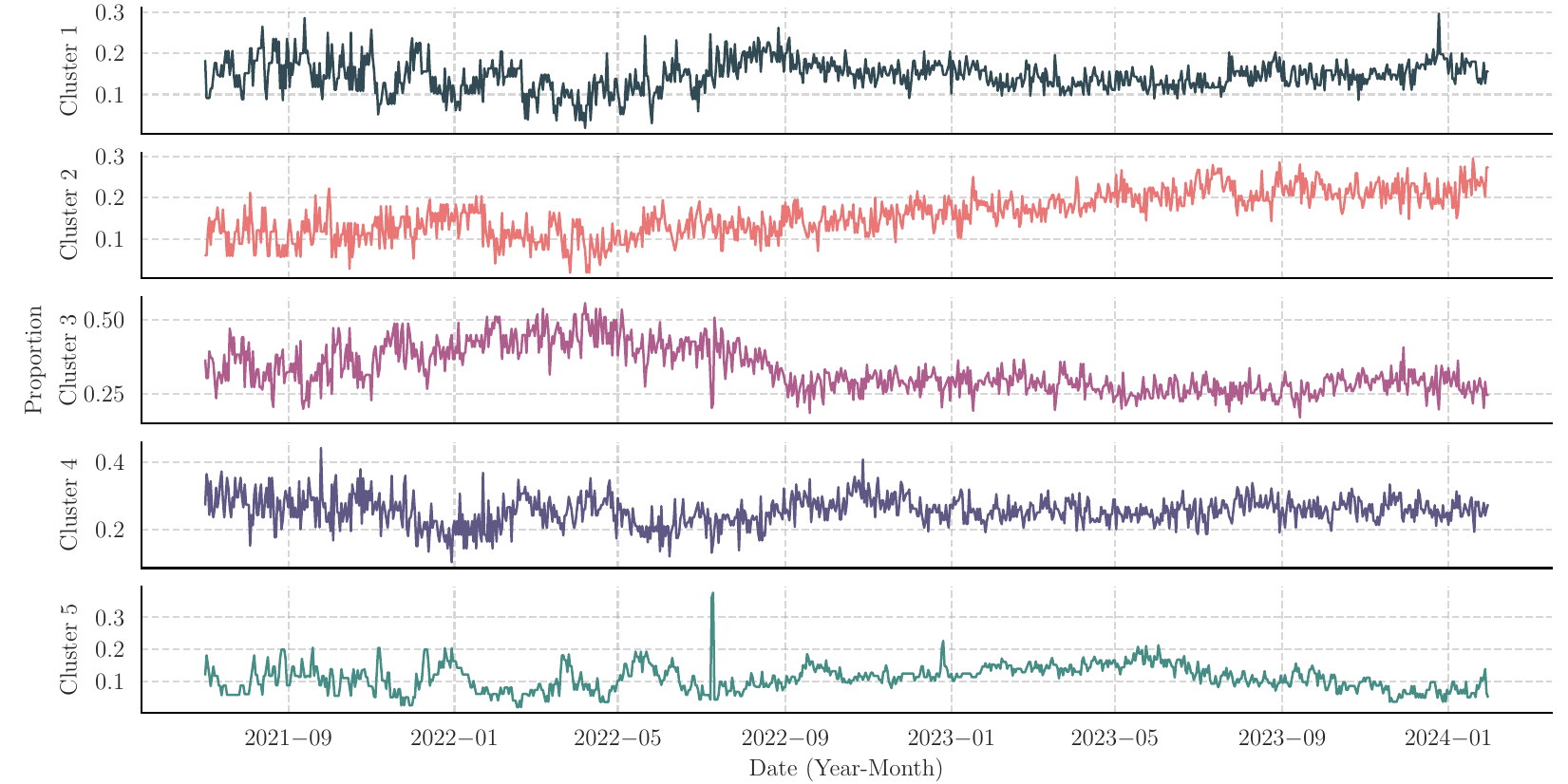}
    \caption{\textbf{Cluster proportions.} Here, for each day in the dataset we show the proportion of days clustered with each label.}
    \label{fig:clutser_proportions}
\end{figure}

In Figure \ref{fig:clutser_proportions}, we present the proportion of participants with a given cluster label for each day in the dataset. Given a semantic understanding of the different days that each cluster represents, we can visualise the behaviour of the participant cohort as a whole. We observe that cluster $2$ generally increases in proportion over the length of the dataset, whilst cluster $3$ seems to rise and fall in proportion between 2021-09 and 2022-09.

\subsection{Vector search over days}
\label{sec:vector_search}

\begin{figure}[ht]
    \centering
    \includegraphics[width=1\linewidth]{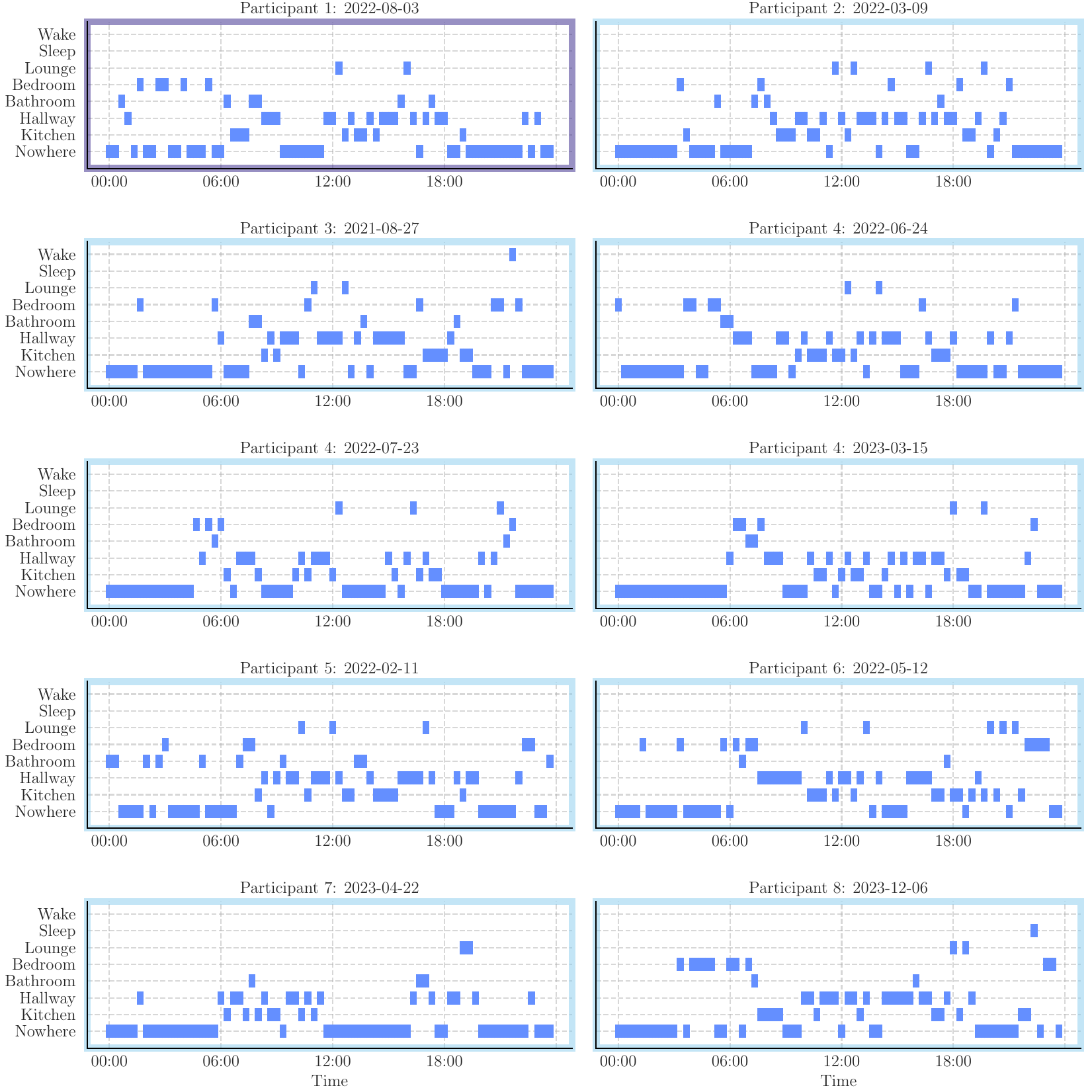}
    \caption{\textbf{Vector search over days.} Given a day of data (shown in purple in the top left), we present the $9$ most similar other days as measured by cosine similarity (shown in blue).}
    \label{fig:vector_search_days}
\end{figure}

To present the use of our method for finding similar days of activity, in Figure \ref{fig:vector_search_days} we show a day of data (chosen at random) and the $9$ most similar days as measured by cosine similarity. These days show similar aspects, and all have large levels of Kitchen and Hallway activity during the day, with small amounts of Lounge, Bedroom, and Bathroom activity.

\subsection{Similarity in days with urinary tract infection labels}
\label{sec:uti_similarity_information}

Here, we will clarify the methods used to calculate the average cosine similarity between days positively labelled with urinary tract infections (UTIs), and those negatively labelled.

We firstly identify the participants and dates associated with a UTI label. We then keep only the participants with at least two positive UTI labels and one negative UTI label. This is done to ensure that for a given participant, we can calculate the similarity between the positive labels (excluding the cosine similarity between the same days) and the similarity between the positively labelled days and the negatively labelled days. After this, we are left with $77$ positively and $77$ negatively labelled days produced by $14$ participants. For each of these participants, we calculate the average cosine similarity between positive days of UTI and the average cosine similarity between the positive and negative days of UTI. We then find the mean and standard deviation of these averages to produce the final values given in Section \ref{sec:experiments}.

\end{document}

%% file: math_commands.tex
\usepackage{amsmath,amsfonts,bm}

\def\eqref#1{equation~\ref{#1}}

\def\1{\bm{1}}

\DeclareMathAlphabet{\mathsfit}{\encodingdefault}{\sfdefault}{m}{sl}
\SetMathAlphabet{\mathsfit}{bold}{\encodingdefault}{\sfdefault}{bx}{n}

